\documentclass[sigconf]{acmart}
 \usepackage{multirow}
 \usepackage{balance}

\AtBeginDocument{%
  \providecommand\BibTeX{{%
    \normalfont B\kern-0.5em{\scshape i\kern-0.25em b}\kern-0.8em\TeX}}}

\copyrightyear{2021} 
\acmYear{2021} 
\setcopyright{acmcopyright}
\acmConference[MM '21]{Proceedings of the 29th ACM
International Conference on Multimedia}{October 20--24, 2021}{Virtual Event, China}
\acmBooktitle{Proceedings of the 29th ACM International Conference on Multimedia
(MM '21), October 20--24, 2021, Virtual Event, China}
\acmPrice{15.00}
\acmDOI{10.1145/3474085.3475618}
\acmISBN{978-1-4503-8651-7/21/10}

\acmSubmissionID{mfp2474}

\begin{document}
\fancyhead{}

\title{Context-Aware Selective Label Smoothing for Calibrating Sequence Recognition Model}

\author{Shuangping Huang$^{1,2}$,\quad Yu Luo$^1$,\quad Zhenzhou Zhuang$^{1}$,\quad Jin-Gang Yu$^{1,2}$*,\quad Mengchao He$^{3}$,\quad Yongpan Wang$^{4}$}

\makeatletter
\def\authornotetext#1{
\if@ACM@anonymous\else
    \g@addto@macro\@authornotes{
    \stepcounter{footnote}\footnotetext{#1}}
\fi}
\makeatother
\authornotetext{Corresponding author.}

\affiliation{
 \institution{\textsuperscript{\rm 1}South China University of Technology, Guangzhou, China}
 \institution{\textsuperscript{\rm 2}Pazhou Laboratory, Guangzhou, China}
 \institution{\textsuperscript{\rm 3}DAMO Academy, Alibaba Group, Hangzhou, China}
 \institution{\textsuperscript{\rm 4}Alibaba Group, Hangzhou, China}
 }
\email{{eehsp,jingangyu}@scut.edu.cn, luoyurl@126.com, zhenzhouzhuang@foxmail.com, mengchao.hmc@alibaba-inc.com, yongpan@taobao.com}

\def\authors{Shuangping Huang, Yu Luo, Zhenzhou Zhuang, Jin-Gang Yu, Mengchao He, Yongpan Wang}

\renewcommand{\shortauthors}{Huang et al.}


\begin{abstract}
Despite the success of deep neural network (DNN) on sequential data (i.e., scene text and speech) recognition, it suffers from the over-confidence problem mainly due to overfitting in training with the cross-entropy loss, which may make the decision-making less reliable. Confidence calibration has been recently proposed as one effective solution to this problem. Nevertheless, the majority of existing confidence calibration methods aims at non-sequential data, which is limited if directly applied to sequential data since the intrinsic contextual dependency in sequences or the class-specific statistical prior is seldom exploited. To the end, we propose a Context-Aware Selective Label Smoothing (CASLS) method for calibrating sequential data. The proposed CASLS fully leverages the contextual dependency in sequences to construct confusion matrices of contextual prediction statistics over different classes. Class-specific error rates are then used to adjust the weights of smoothing strength in order to achieve adaptive calibration. Experimental results on sequence recognition tasks, including scene text recognition and speech recognition, demonstrate that our method can achieve the state-of-the-art performance.
\end{abstract}




\maketitle

\section{Introduction}

Benefited from the remarkable fitting capacity of deep learning models, the performance of sequential data recognition has been largely boosted recently, which consequently enables their deployment in plenty of user-facing applications, such as clinical text recognition \cite{KimL20}, autonomous driving \cite{BaoYK20} and speech recognition \cite{5749278}. However, these deep models, most typically trained with the cross-entropy loss function, tend to assign a high probability even for a wrong prediction, which is the so-called over-confidence problem of deep sequential models.

\begin{figure}[tp]
  \centering
  \includegraphics[width=0.5\linewidth]{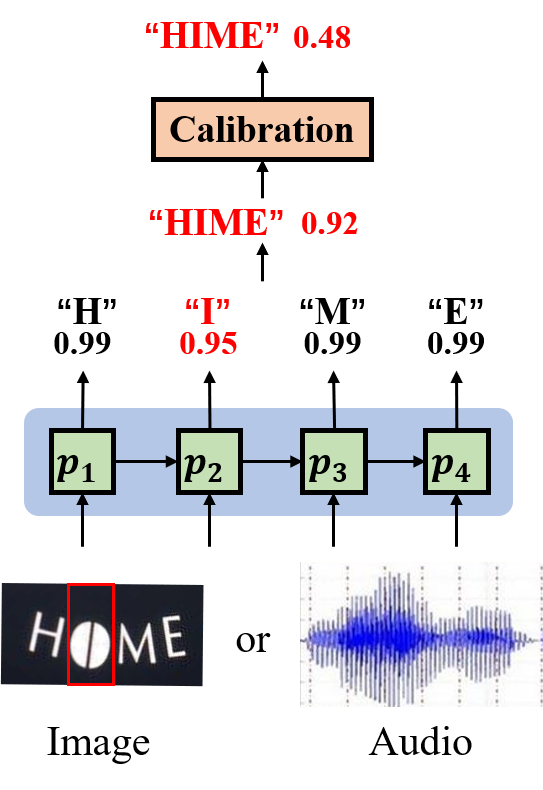}
  \caption{The prediction of each token in a sequence is not independent. Therefore, we focus on calibrating the confidence for the entire sequence instead of for individual token.}
\label{figure1}
\end{figure}

Confidence calibration has been proposed as one popular solution to the over-confidence problem. Typical methods in machine learning, including Platt scaling\cite{ZadroznyE01}, isotonic regression\cite{NaeiniCH15} and histogram binning\cite{2000Probabilistic}, inspire the confidence calibration of deep models \cite{GuoPSW17, MullerKH19}. Unfortunately, the majority of the existing methods mainly focus on non-sequential data, which cannot be trivially adapted to the calibration of sequential data. One key reason is that the task of sequence recognition involves a temporal process of predicting each token in order, where the contextual dependency among tokens is of critical importance (see in Fig. \ref{figure1}). Therefore, simply conducting uniform calibration on each token independently, unaware of inter-token dependency, is limited for sequential data.

There have already been a couple of previous attempts on confidence calibration of sequence recognition. Some authors considered the sequential characteristics and attempted to incorporate them into the sequential framework\cite{KuleshovL15}. The varying sequential length is utilized as the setup basis of temperature values based on the temperature scaling technique \cite{GuoPSW17,abs-2002-02644, abs-2012-12643}. Attention scores calculated during the sequence modeling are also utilized as the weights of temperature values \cite{abs-1903-00802}. In natural language processing (NLP), some measurements of sequence such as N-gram and BLEU\cite{PapineniRWZ02} are exploited to guide the confidence allocation of correct and wrong predictions, to avoid the over-confidence typically caused by the cross-entropy loss \cite{LukasikJMKBYK20, GaoWHYN20, abs-2004-03437,SongZWH20}. Although the mentioned sequential properties are taken into account in these works, they still have two main limitations: 1) The influence of contextual dependency is seldom explored, and specialized manners of contextual feature extraction in NLP also make it hard to be generalized to sequential recognition tasks. 2) These methods ignore the uneven class distribution of predictions globally, while only considering the pairwise relationship between the prediction and the ground truth locally.

In this paper, we propose a Context-Aware Selective Label Smoothing method to implement a sequence-level calibration for sequential recognition models. To fully exploit the contextual dependency underlying the sequence, we construct confusion matrices for each class to represent the prediction distribution according to the contextual relation. Statistics are derived from a support set, which resembles the data distribution of the training set and thus is served as a reference for prediction distribution. Specifically, the probability distribution of labels belonging to different classes is reallocated based on the corresponding error rate in the confusion matrix, wherein error-prone labels are assigned with stronger smoothing strength. In this way, label smoothing on sequence can be implemented adaptively by considering contextual dependency and unbalanced prediction classes simultaneously. 
To evaluate the effectiveness of the proposed method, we conduct experiments on two sequence recognition tasks, including scene text recognition and speech recognition. Experimental results demonstrate that our method is superior to the state-of-the-art calibration methods, and also can be well generalized on sequence recognition tasks.

The main contributions of this paper are summarized as:

\begin{enumerate}

\item We propose a novel sequence-level confidence calibration method to adaptively calibrate the over-confident model.   
\item The contextual dependency of sequence is fully explored to construct confusion matrices of contextual token prediction for each class, which is utilized as the weight basis for adaptive label smoothing. 
\item We demonstrate that our proposed method can generalize well on sequence recognition tasks to achieve calibrated results with a certain improvement on recognition accuracy.

\end{enumerate}


\section{Related Work}

\subsection{Calibration on Non-sequential Data} 

The well-established calibration methods for non-sequential data related tasks, such as image classification, are borrowed from machine learning techniques\cite{2000Probabilistic, ZadroznyE01, ZadroznyE02, NaeiniCH15}. With the prosperity of deep learning, the calibration of deep neural networks has recently attracted a lot of research interests in the literature. Post-processing based calibration methods learn a regression function based on a small held-out dataset to adjust the output confidence value without retraining the neural network from scratch and influencing the accuracy. Temperature scaling is the most commonly used method proposed by Guo et al. \cite{GuoPSW17}. It is a variant version of Platt scaling through introducing a temperature parameter \cite{2000Probabilistic} to adaptively calibrate the models. Ji et al. \cite{JiJYKS19} further implement temperature scaling in a bin-wise setting to improve the performance via denser prediction interval. However, as the confidences of all the predictions are calibrated with fixed parameters, it may lead to the underestimation of the predictions. Additional methods calibrate the confidence through optimizing the loss function in the training process. Label smoothing \cite{SzegedyVISW16} is proposed as a regularization technique to soften the one-hot encoding distribution to calibrate the model \cite{MullerKH19}. Kumar et al. \cite{KumarSJ18} and Mukhoti et al. \cite{MukhotiKSGTD20} replace the general cross entropy loss with the distance between accuracy and confidence score and focal loss \cite{LinGGHD17}, respectively. The decent results are achieved on non-sequential data, however, these methods are hard to be directly applied to the sequence. Simply conducting a uniform calibration across all the tokens of a sequence is too aggressive to produce good calibration performance.

\subsection{Calibration on Sequential Data}

Instead of trivially applying non-sequential calibration methods on each token, some methods attempt to exploit the intrinsic characteristics to facilitate sequence-level calibration.  Kuleshov and Liang \cite{KuleshovL15} firstly consider confidence calibration in the context of structure prediction problem, which demonstrates that utilizing structural features will produce better calibration performance on structured prediction such as sequence. Recent methods for sequential data can be categorized into two directions. Based on temperature scaling \cite{GuoPSW17}, Leathart et al. propose a sequential length-related calibration method \cite{abs-2002-02644}. As calibration error varies with sequential length, different temperature values are set for calibration. Ding et al. obtain the adaptive temperature values for different pixels or voxels for semantic segmentation \cite{abs-2008-05105}. Other methods are inspired by label smoothing \cite{SzegedyVISW16}. In natural language processing, Elbayad et al. complete token-level smoothing with sequence-level smoothing methods. The nearby target sequence is sampled according to the BLEU score \cite{PapineniRWZ02} to conduct sequence-level loss smoothing \cite{BesacierEV18}. Lukasik et al. utilize n-gram overlapping with the target sequence in the label smoothing method to guarantee the semantic similarity of translated sequence \cite{LukasikJMKBYK20}. Song et al. smooth the loss function according to the occurrence of the n-gram context in candidate word set as the weight \cite{SongZWH20}. Although better calibration performance could be achieved on sequential data, few methods consider the uneven distribution of dataset to flexibly calibrate the prediction of different labels, and additionally, the potential contextual information is not fully explored to calibrate on the recognition of sequential data tasks.

\section{Preliminaries}

\subsection{Sequential Confidence}

In sequential data recognition tasks, we aim to assign an $L$-length sequential label vector $Y=\left\{y_1, y_2, \cdots, y_L\right\}$ to the input $x$. For example, in scene text recognition, $x$ is an image containing a text instance and $Y$ is the sequence of character tokens. And in speech recognition, $x$ is a sound waveform and $Y$ is the sequence of phoneme tokens. For sequence prediction, the attention-based model consisting of RNN cells is commonly adopted. The token is predicted based on the information of previous states and the proceeding token. Assuming the set of class labels being $\{1,...,K\}$, the predicted token $\hat{y}_t$ is obtained by maximizing the conditional probability as below:

\begin{equation}
    \hat{y}_t= \mathop{\arg \max}_{1\leq k \leq K} p_t(y_t = k|x,y_{<t})
\end{equation}
where the $y_{<t}$ represents the previous tokens before the time step $t$. The token in the sequence is predicted in a temporal manner. After decoding at each time step, the sequence consisting of a series of tokens $\hat{Y}=(\hat{y}_1, \hat{y}_2,\cdots, \hat{y}_{L^{'}})$ of length $L^{'}$ can then be predicted.

For the calibration of sequential data, our objective is to optimize the holistic confidence score of sequence $\mathbb{P}(\hat{Y}|x)$, instead of individual token confidence $p_t(y_t=\hat{y}_t|x)$. Probabilistically, the sequential confidence $\mathbb{P}(\hat{Y}|x)$ is a conditional joint probability that can be decomposed as follows:

\begin{align}
\label{eq1}
    \mathbb{P}(\hat{Y}|x)=&\mathbb{P}(\hat{y}_1, \hat{y}_2,\cdots, \hat{y}_{L^{'}}|x) c\\
     = &\mathbb{P}(\hat{y}_1|x) \times \mathbb{P}(\hat{y}_2|x,\hat{y}_1) \times \cdots \times \mathbb{P}(\hat{y}_{L^{'}}|x,\hat{y}_1,\cdots,\hat{y}_{L^{'}-1}) \notag \\
    =& \prod_{t=1}^{L^{'}}{p_t(y_t=\hat{y}_t|x,\hat{y}_{<t})}
\end{align}
where $\hat{y}_{<t}$ represents the previous tokens predicted before time step $t$. As shown in Eq.~\ref{eq1}, the cumulative multiplication of token confidence $p_t(y_t=\hat{y}_t|x,\hat{y}_{<t})$ predicted by a the sequence models, can probabilistically represent the confidence of sequence $\mathbb{P}(\hat{Y}|x)$.

\subsection{ Evaluation of Confidence Calibration}

Calibration requires the confidence score to reflect the true accuracy of prediction. To evaluate whether a sequential model is well calibrated, several common metrics are introduced as follows.

\textbf{Brier Score} \cite{1950Verification} The brier score(BS) is known as the mean square error (MSE). By counting each correctly predicted sequence as one, while each wrong one as zero, the brier score is then defined by:

\begin{equation}
  Brier Score =\frac{1}{N} \sum_{i=1}^N{(\mathbb{I}(\hat{Y}_i = Y_i) - \mathbb{P}(\hat{Y}_i|x))^2},
\end{equation}
where $N$ denotes the total number of samples of testing data,$\mathbb{I}(\cdot)$ is the indicator function, and $\mathbb{P}(\hat{Y}_i|x)$ denotes the confidence of the $i$-th sequence.
    
\textbf{Expected Calibration Error (ECE)} \cite{NaeiniCH15} The ECE evaluates the calibration performance through measuring the absolute distance between the confidence and accuracy in a bin-wise manner. Firstly, the probability space [0,1] is split into $M$ bins, and the test samples are also divided into different bins according to the corresponding confidence scores. The accuracy and the average confidence of the $m$-th bin are calculated as follows:

\begin{equation}
  Acc(B_m) = \frac{1}{|B_m|}\sum_{\hat{Y}_i \in B_m}{\mathbb{I}(\hat{Y}_i = Y_i)},
\end{equation}

\begin{equation}
  Conf(B_m) = \frac{1}{\left\vert B_m \right\vert}\sum_{ \hat{Y}_i\in B_m }{\mathbb{P}(\hat{Y}_i|x)},
\end{equation}
where $B_m$ and $\left\vert B_m \right\vert$ are the test samples and the number of samples in
the $m$-th bin, respectively. And the ECE is formally defined as:

\begin{equation}
ECE = \sum_{m=1}^M{\frac{\left\vert B_m \right\vert}{N}\left\vert Acc(B_m) - Conf(B_m) \right\vert},
\end{equation}

For a well-calibrated model, the ECE value should be close to 0 with the accuracy is equal to the confidence score.

\textbf{Reliability Diagram} \cite{2987588} The reliability diagram is the plot of $Acc\left(B_m \right)$ versus $Conf\left(B_m \right)$. As is shown in Fig. \ref{figure5} of Sec. \ref{Experiment}, the solid concave curve in the reliability diagram represents the over-confidence state, while the convex one represents the confidence is less than the expected accuracy. If the model is well calibrated, the solid line should be aligned with the dotted diagonal line. 

\begin{figure}[tp]
  \centering
  \includegraphics[width=\linewidth]{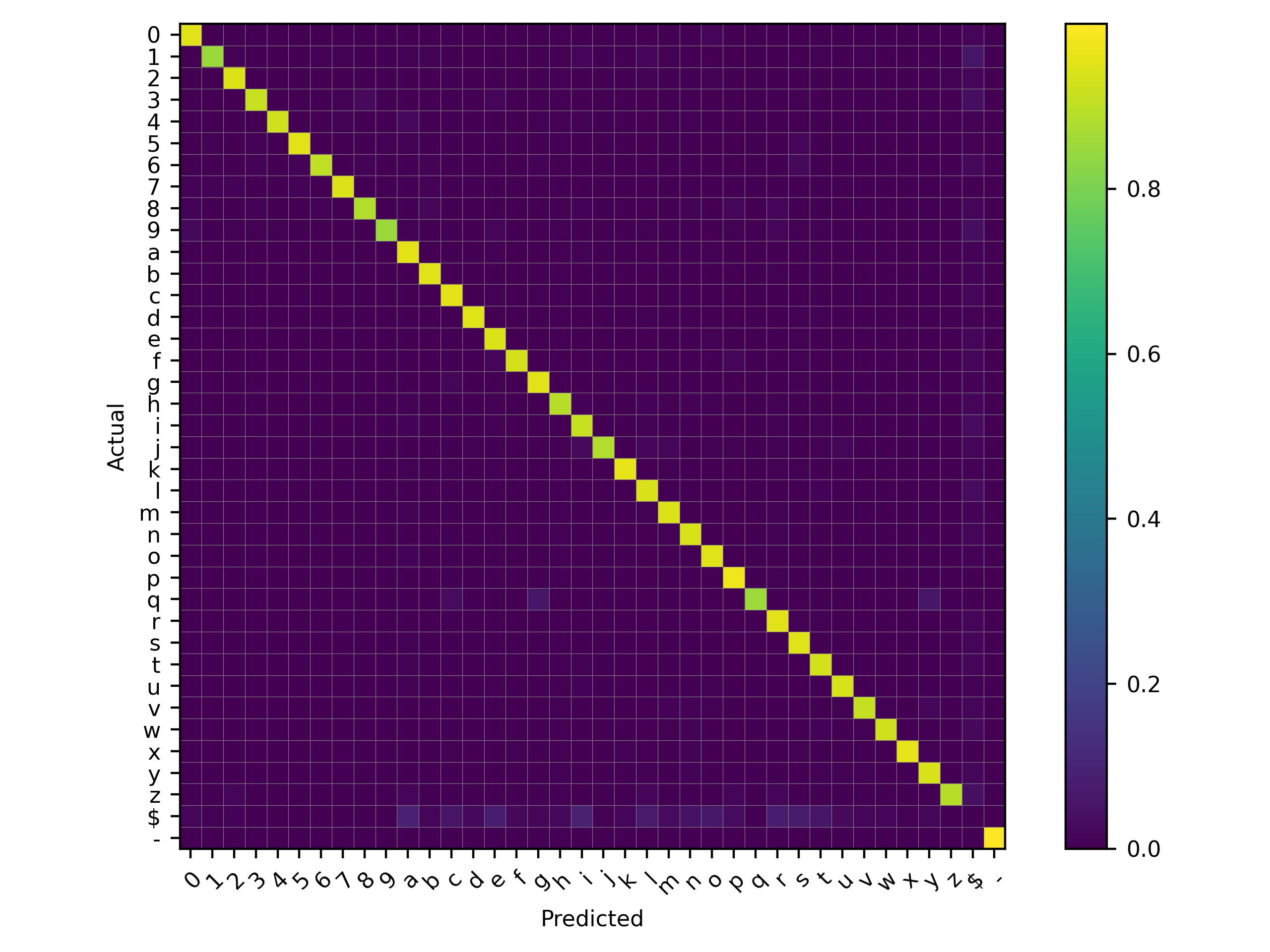}
  \caption{The confusion matrix of the support dataset on scene text recognition. The predicted error frequency of character ``j'' is larger than character ``a''.}
\label{figure2}
\end{figure}

\begin{figure*}[ht]
\begin{center}
\includegraphics[width=1.0\linewidth]{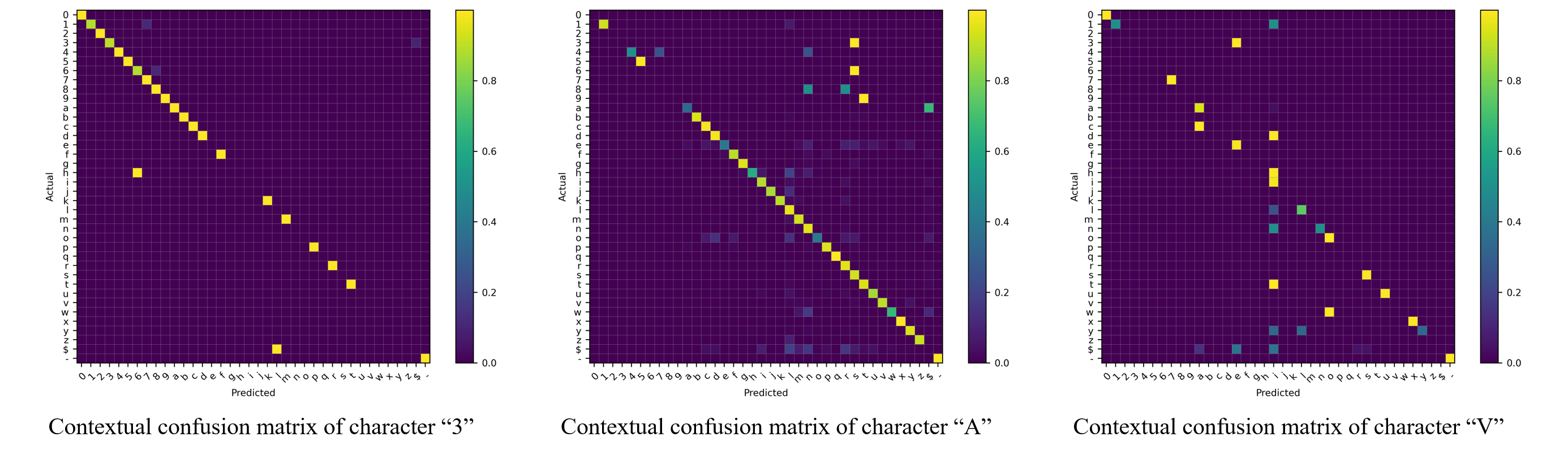}
\end{center}
   \caption{The example contextual confusion matrices of different tokens on scene text recognition.}
\label{figure3}
\end{figure*}

\subsection{Standard Label Smoothing}
\label{Standard Label Smoothing} 
As the model training is associated with cross entropy loss using one-hot encoding, the large distance of 0-1 distribution is prone to make the model become over-confident for the prediction, which is undesirable for decision-making. Thus, the standard label smoothing (LS) is proposed as a common regularization technique for deep model \cite{SzegedyVISW16}. It smooths the one-hot distribution with a 
hyper-parameter $\alpha$ to obtain a soft distribution $\hat{q}_t(y_t=k)$ of every token $y_t$:

\begin{equation}
\hat{q}_t(y_t=k) = \left\{ {\begin{array}{c}
     1-\alpha, \quad if \, y_t=k \\
     \frac{\alpha}{K-1}, \quad if \, y_t \neq k\\ 
\end{array}} \right.
\end{equation}

Although LS is proved to be helpful to alleviate the over-confident problem\cite{MullerKH19} by smoothing confidence $1-\alpha$ in soft distribution, imposing the same smoothing strength on the label of all the tokens indiscriminately is inapplicable for calibrating sequence. According to the Eq. \ref{eq1}, even if all the tokens are correctly predicted, the sequential confidence is calculated as $\mathbb{P}(\hat{Y}|x)\approx (1-\alpha)^{L^{'}}$, which conversely leads to an exponential reduction on confidence.

\section{Method}
\label{method}


Considering the uneven prediction distribution of different classes, we first design a selective label smoothing (SLS) method to conduct an adaptive calibration based on the confusion matrix of misprediction statistics (subsection \ref{SLS}). Furthermore, we exploit the contextual dependency in the sequence and propose a context-aware selective label smoothing method based on SLS (subsection \ref{CASLS}).

\subsection{Selective Label Smoothing}
\label{SLS}
The standard label smoothing method smooths the one-hot distribution of all classes evenly, while ignoring the differences between different character categories. It is a common phenomenon that the predicted error distribution varies from class to class. As shown in Fig. \ref{figure2}, the predicted error rate of character “j” is large than character “a”. It is reasonable to reduce the prediction confidence of error-prone character "j" by standard label smoothing. However, reducing the prediction confidence of non-error-prone character "a" is unreasonable. In this case, we focus on the frequent error-prone classes to implement an adaptive label smoothing. For the classes with few or even zero errors, it is needless to conduct label smoothing. 

To obtain the statistical error-prone classes, we employ a support datasets that resembles the data distribution of the training set as the reference prediction distribution. And we construct a confusion matrix $C_m=(c_{i,j})\in R^{K \times K}$ to quantitatively represent the prediction of $K$ classes, where $c_{i,j}$ denotes the number of the elements of the $i$-th class but are recognized as the $j$-th class. 
Fig. \ref{figure2} is an example confusion matrix of the support dataset on scene text recognition task. Outsides the correct prediction distributed on the diagonal line, the misprediction distribution of some classes is higher than that of other classes. 

Here, we set a threshold to select those classes with a high error rate. And the error-prone set $E$ can be defined:

\begin{equation}
    E=\left\{ k \mid 1 - \frac{c_{k,k}}{\sum_j{c_{k,j}}} > T , k = 1,2,\cdots, K \right\}
\end{equation}
where $T$ denotes the probability threshold of the error-prone tokens.

And the selective smoothed label distribution $\hat{q}(y_t=k)$ can be represented as follows:

\begin{equation}
\hat{q}(y_t=k) = \left\{ {\begin{array}{c}
     1 , \quad if \, y_t=k \, and \, y_t \notin E \\
     1-\alpha, \quad if \, y_t=k \, and \, y_t \in E \\
     0 , \quad if \, k \neq y_t \, and \, y_t \notin E \\
     \alpha \frac{c_{y_t,k}}{\sum_j{c_{y_t,j}}}, \quad if \, y_t \neq k \, and \, y_t \in E\\
\end{array}} \right.
\end{equation}

If the token $y_t$ falls into the error-prone set $E$, a corresponding smoothing strategy will be conducted on the token. The error rate $\frac{c_{y_t,k}}{\sum_j{c_{y_t,j}}}$ is used as the weight of smoothing strength $\alpha$, when the prediction $k\neq y_t$. 

\subsection{Context-aware Selective Label Smoothing}
\label{CASLS}

Considering the contextual dependency existing in the sequential recognition model, the prediction errors distribution of every token depends on the different previous token. The error-prone set is related to the class of the previous token. Thus, we further propose a context-aware selective label smoothing method. We analyze the prediction error distribution under the condition of previous token of different classes, and similarly represent it in the form of a confusion matrix. Then, We obtain $K$ confusion matrices $\left\{C_k \mid C_k=(c_{k,i,j})\in R^{K \times K}, k = 1, 2, \cdots, K\right\}$. The $C_k$ is the confusion matrix of predicted token, when the preceding token belongs to the $k$-th class. We name it as context confusion matrix. And $c_{k, i, j}$ represents the number of current token belonging to the $i$-th class that is recognized as the $j$-th class for the $k$-th class of the preceding token. 

Fig. \ref{figure3} is the example context confusion matrices of different classes on scene text recognition.  Errors rarely occur on the character tokens, when the preceding token is a digit ``3''. There are few prediction errors for most character tokens when the preceding token is an ``A''. But errors often occur on the character ``C'' when the preceding token is an ``V''. It can be observed that it is rational to conduct a context related calibration since the misprediction distribution varies from class to class.

Similar to the process of SLS, totally $K$ error-prone sets $E_k$ are obtained:

\begin{equation}
    E_k =\left\{ i \mid 1 - \frac{c_{k,i,i}}{\sum_j{c_{k,i,j}}} > T , i = 1,2,\cdots, K \right\},
\end{equation}
where $E_k$ denote the error-prone token set of the $k$-th class.

Therefore, the conditional probability distribution of context-aware selective label smoothing is correlated with the class of the preceding token $y_{t-1}$. The soft-target distribution $\hat{q}(y_t=k \mid y_{t-1})$ is computed as:

\begin{equation}
\hat{q}(y_t=k \mid y_{t-1}) = \left\{ {\begin{array}{c}
     1 , \quad if \, y_t=k \, and \, y_t \notin E_{y_{t-1}} \\
     1-\alpha, \quad if \, y_t=k \, and \, y_t \in E_{y_{t-1}} \\
     0 , \quad if \, y_t \neq k \, and \, y_t \notin E_{y_{t-1}} \\
     \alpha \frac{c_{y_{t-1},y_t,k}}{\sum_j{c_{y_{t-1},y_t,j}}}, \quad if \, y_t \neq k \, and \, y_t \in E_{y_{t-1}}\\
\end{array}} \right.
\end{equation}

When the previous token is $y_{t-1}$, the distribution $\hat{q}(y_t=k \mid y_{t-1})$ will adaptively select the error-prone set $E_{y_{t-1}}$. If the current token $y_t$ is in the error-prone set $E_{y_{t-1}}$, a corresponding smoothing strategy will be conducted on the token $y_t$.

\begin{equation}
    Loss = -\frac{1}{L}\sum_{t=1}^L{\sum_{k=1}^K{\hat{q}(y_t=k \mid y_{t-1})log(P(\hat{y}_t = k))}}.
\end{equation}
The corresponding loss function will be redefined in the training process.

\subsection{Sequence Alignment Strategy}

In sequence recognition task, the existence of the omissive or redundant token in the prediction sequence sometimes leads to the misalignment between it and the reference sequence. Such misalignment results in two problems. Firstly, since one token is missing in the sequence prediction, the sequence confidence would be inaccurate without the attendance of the probability of the omissive token. The similar explanation is applied for redundant tokens. And the other problem is that the missing or redundant token makes it confusing to determine the alignment relationship between the ground truth token and the corresponding predicted token, which also influences the construction of confusion matrix.

\begin{figure}[ht]
\begin{center}
\includegraphics[width=\linewidth]{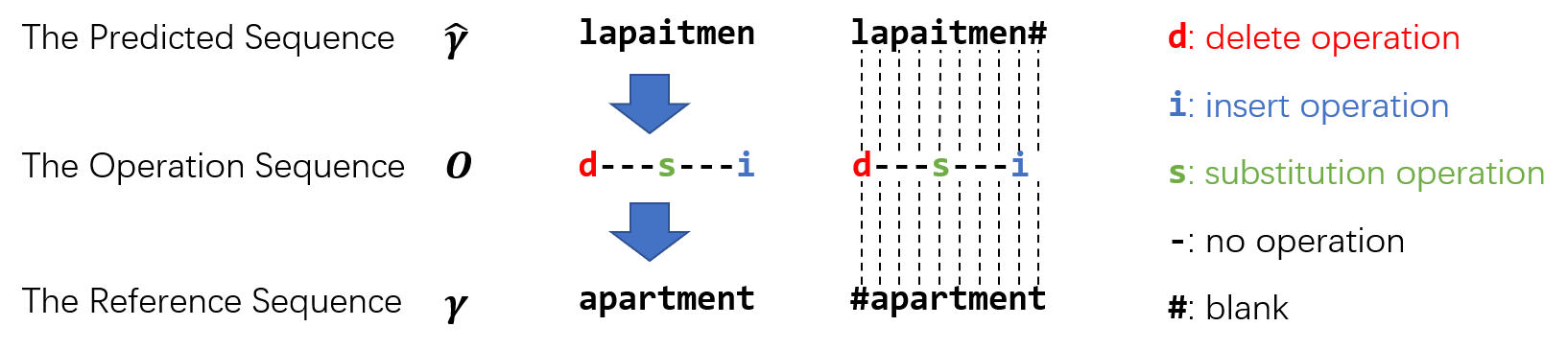}
\end{center}
   \caption{The illustration of the proposed sequence alignment strategy.}
\label{figure4}
\end{figure}

To handle the misalignment problem, we introduce a ``blank'' class to fill the place of the missing token, and track the alignment between prediction and ground truth sequence along an operator sequence of edit distance. For example, as is shown in Fig. \ref{figure4}, the misalignment between predicted and reference sequence is caused by the redundant character ``l'' and the missing character ``t'', which correspond to the deletion and insertion operation in edit distance, respectively. By tracking the flag of edit distance (deletion and insertion), we complement the ``blank'' space in the corresponding position to realize the alignment.

\subsection{Adaptive Smoothing Strength}

As is analyzed in subsection \ref{Standard Label Smoothing}, the sequential confidence will become quite small due to the cumulative effect of tokens confidence smoothed via the same hyper-parameter $\alpha$. In order to solve the above-mentioned less confident problem of sequence, a trick is introduced to label smoothing. We adaptively adjust the value of hyper-parameter $\alpha$ according to the length $L$ of sequence. For each sequence, $\alpha=1-\sqrt[L]{1-\alpha^{'}}$, which compensates for the negative impact of less confidence caused by the cumulative calculation.

 \begin{figure*}[ht]
\begin{center}
\includegraphics[width=1.0\linewidth]{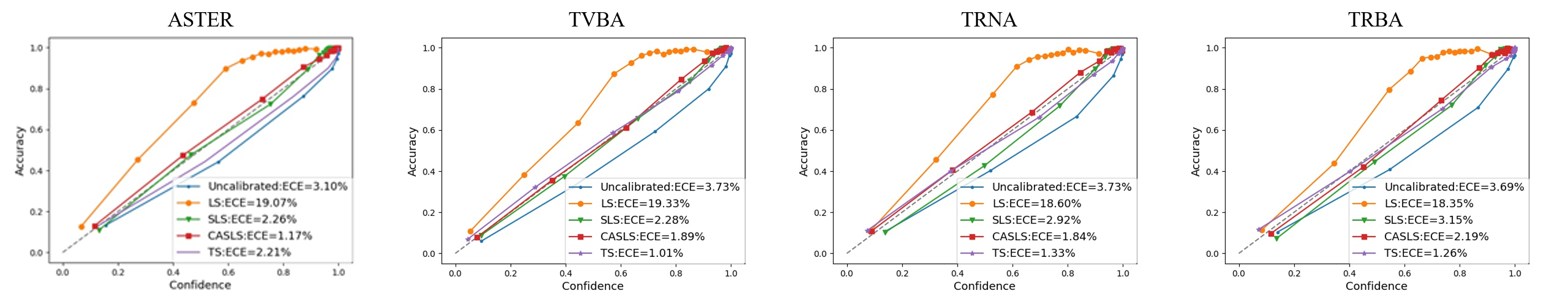}
\end{center}
   \caption{The reliability diagrams (with $B_m$=15 bins) of the uncalibrated models and the models calibrated with LS, SLS and CASLS on STR task.}
\label{figure5}
\end{figure*}

\begin{table*}[ht]
\caption{The calibration results comparison of LS, TS, SLS and CASLS on STR task. The three metrics: accuracy(Acc), ECE and brier score(BS) are listed.}
\label{table1}
\resizebox{\textwidth}{21mm}{
\begin{tabular}{ccccccccccccc}
\hline
\multirow{2}{*}{Method} & \multicolumn{3}{c}{ASTER}    & \multicolumn{3}{c}{TVBA}   & \multicolumn{3}{c}{TRNA}       & \multicolumn{3}{c}{TRBA}    \\ \cline{2-13} 
     & Acc(\%) $\uparrow$ & ECE(\%) $\downarrow$    & BS(\%) $\downarrow$  &
     Acc(\%) $\uparrow$ & ECE(\%) $\downarrow$   & BS(\%)$\downarrow$  & 
    Acc(\%) $\uparrow$ & ECE(\%) $\downarrow$  & BS(\%) $\downarrow$  &
    Acc(\%) $\uparrow$ & ECE(\%) $\downarrow$ & BS(\%)  $\downarrow$    \\ \hline
Uncalibrated      & 87.33        & \begin{tabular}[c]{@{}c@{}}3.10\\ (+0\%)\end{tabular} &   
\begin{tabular}[c]{@{}c@{}}5.24\\ (+0\%)\end{tabular}
& 83.83        & \begin{tabular}[c]{@{}c@{}}3.73\\ (+0\%)\end{tabular}   &
\begin{tabular}[c]{@{}c@{}}6.01\\ (+0\%)\end{tabular}
& 85.91        & \begin{tabular}[c]{@{}c@{}}3.73\\ (+0\%)\end{tabular}   &
\begin{tabular}[c]{@{}c@{}}5.73\\ (+0\%)\end{tabular}
& 86.43        & \begin{tabular}[c]{@{}c@{}}3.69\\ (+0\%)\end{tabular}  &
\begin{tabular}[c]{@{}c@{}}5.52\\ (+0\%)\end{tabular}\\ \hline
LS        & 86.32        & \begin{tabular}[c]{@{}c@{}}19.07\\ (+515\%)\end{tabular} &
\begin{tabular}[c]{@{}c@{}}10.04\\ (+92\%)\end{tabular}
& 84.87        & \begin{tabular}[c]{@{}c@{}}19.33\\ (+418\%)\end{tabular} &
\begin{tabular}[c]{@{}c@{}}10.41\\ (+42\%)\end{tabular}
& 86.23        & \begin{tabular}[c]{@{}c@{}}18.60\\ (+399\%)\end{tabular} &
\begin{tabular}[c]{@{}c@{}}9.92\\ (+73\%)\end{tabular}
& 86.29        & \begin{tabular}[c]{@{}c@{}}18.35\\ (+397\%)\end{tabular} &
\begin{tabular}[c]{@{}c@{}}9.73\\ (+43\%)\end{tabular}\\ \hline

TS      & 87.33        & \begin{tabular}[c]{@{}c@{}}2.21\\ (-29\%)\end{tabular} &   
\begin{tabular}[c]{@{}c@{}}4.97\\ (-5\%)\end{tabular}
& 83.83        & \begin{tabular}[c]{@{}c@{}}1.01\\ (-72\%)\end{tabular}   &
\begin{tabular}[c]{@{}c@{}}5.77\\ (-4\%)\end{tabular}
& 85.91        & \begin{tabular}[c]{@{}c@{}}1.33\\ (-64\%)\end{tabular}   &
\begin{tabular}[c]{@{}c@{}}5.40\\ (-6\%)\end{tabular}
& 86.43        & \begin{tabular}[c]{@{}c@{}}1.26\\ (-66\%)\end{tabular}  &
\begin{tabular}[c]{@{}c@{}}5.21\\ (-6\%)\end{tabular}\\ \hline

SLS       & 87.09        & \begin{tabular}[c]{@{}c@{}}2.26\\ (-27\%)\end{tabular}  &
\begin{tabular}[c]{@{}c@{}}5.08\\ (-3\%)\end{tabular}
& 85.14        & \begin{tabular}[c]{@{}c@{}}2.28\\ (-39\%)\end{tabular}  &
\begin{tabular}[c]{@{}c@{}}5.64\\ (-6\%)\end{tabular}
& 86.49        & \begin{tabular}[c]{@{}c@{}}2.92\\ (-22\%)\end{tabular}  &
\begin{tabular}[c]{@{}c@{}}5.30\\ (-8\%)\end{tabular}
& 86.56        & \begin{tabular}[c]{@{}c@{}}3.15\\ (-15\%)\end{tabular}   &
\begin{tabular}[c]{@{}c@{}}5.02\\ (-9\%)\end{tabular}\\ \hline
CASLS     & 87.31        & \begin{tabular}[c]{@{}c@{}}1.17\\ (-62\%)\end{tabular}   &
\begin{tabular}[c]{@{}c@{}}5.18\\ (-1\%)\end{tabular}
& 84.8         & \begin{tabular}[c]{@{}c@{}}1.89\\ (-49\%)\end{tabular}  &
\begin{tabular}[c]{@{}c@{}}5.55\\ (-8\%)\end{tabular}
& 86.11        & \begin{tabular}[c]{@{}c@{}}1.84\\ (-51\%)\end{tabular}  &
\begin{tabular}[c]{@{}c@{}}5.34\\ (-8\%)\end{tabular}
& 86.61        & \begin{tabular}[c]{@{}c@{}}2.19\\ (-41\%)\end{tabular}  &
\begin{tabular}[c]{@{}c@{}}5.09\\ (-8\%)\end{tabular}\\ \hline
\end{tabular}}
\end{table*}


\section{Experiment and Analysis}
\label{Experiment}

In this section, the experimental results are described and discussed. As our method focuses on sequential data, we conduct it on two sequential recognition tasks: scene text recognition (STR) and speech recognition(SR). For each task, the experimental setup including datasets, base models and implementation details are firstly described. Furthermore, we discuss the calibration results under different conditions: uncalibrated model, the model calibrated via the LS, temperature scaling (TS) and the proposed methods.

\subsection{Scene text recognition}
\subsubsection{Experiment setup}

We evaluate the proposed methods on the available benchmark datasets, including four regular datasets: IIIT5K-Words (IIIT5k) \cite{MishraAJ12}, Street View Text (SVT) \cite{WangBB11}, ICDAR 2003 (IC03) \cite{LucasPSTWYANOYMZOWJTWL05}, ICDAR 2013 (IC13)\cite{KaratzasSUIBMMMAH13}, and three irregular datasets: ICDAR 2015 (IC15) \cite{KaratzasGNGBIMN15}, SVT-Perspective (SVTP) \cite{PhanSTT13}, CUTE80 (CUTE) \cite{RisnumawanSCT14}. And the support dataset is the ensemble of the training datasets of IIIT5k, SVT, IC03, IC13 and IC15. The support dataset contains a total of 8539 text instances and resembles the data distribution of training dataset \cite{JaderbergSVZ14, GuptaVZ16}, from which we can get a general prediction distribution.

The experiment is conducted on the state-of-the-art models: ASTER model proposed by Shi et al.\cite{ASTER}, and modular STR framework proposed by Baek et al. \cite{BaekKLPHYOL19}. ASTER is an attention-based model that uses a Spatial Transformer Network \cite{STN} for rectifying oriented or curved text, and a BiLSTM for sequence modeling. For the modular STR framework, we uniformly use the Spatial Transformer Network (T) and attention-based decoder (A). Specifically, the VGG-16 (V) \cite{SimonyanZ14a} and ResNet-34 (R) \cite{HeZRS16} are utilized as backbone networks. And the model with or without a BiLSTM is considered (B/N). We adopt three architectures of different module combinations: TVBA, TRNA and TRBA.

During the calibration, the dimension $K$ of the confusion matrix corresponds to the number of prediction classes (26 characters, 10 digits and 1 blank) of STR task. We obtain confusion matrix and context confusion matrix from the predictions of the uncalibrated model on the support set, and then use SLS and CASLS to fine-tune the uncalibrated model. Here, we set up the threshold of the error-prone $T$=0.5, the smoothing parameter $\alpha^{'}$=0.05, and the interval width of each bin $B_m$=15.

\subsubsection{Results and Discussion}

As is shown in Fig. \ref{figure5}, the blue curves below the diagonal line represent that all the uncalibrated STR models tend to be over-confident. As we analyzed before, the models trained with the LS(orange curves in Fig. \ref{figure5}) will become less confident instead, the curves is obviously convex in the high-confidence intervals. For all the calibration methods, SLS (green curves), TS (purple curves) and CASLS (red curves) approach are superior to the LS approach, among which the curves of TS and CASLS almost coincide with the diagonal line and can achieve the best calibration performance.

Table \ref{table1} shows the quantitative calibration results of LS, TS and the proposed SLS and CASLS methods. The accuracy, ECE and brier score(BS) are adopted as the measurements. In terms of ECE and BS, comparing with LS that widens the distance between accuracy and confidence, SLS and CASLS can achieve a good calibration performance on sequence through considering the uneven prediction distribution of different classes. And it is noted that, due to the utilization of contextual dependency, CASLS further outperforms SLS. And TS outperforms both of the proposed methods on TVBA, TRNA and TRBA. However, TS will not influence the accuracy, while the proposed methods can improve the accuracy based on the base model, and about 1.3\% improvement is achieved on TVBA. 

\begin{figure}[ht]
  \centering
  \includegraphics[width=\linewidth]{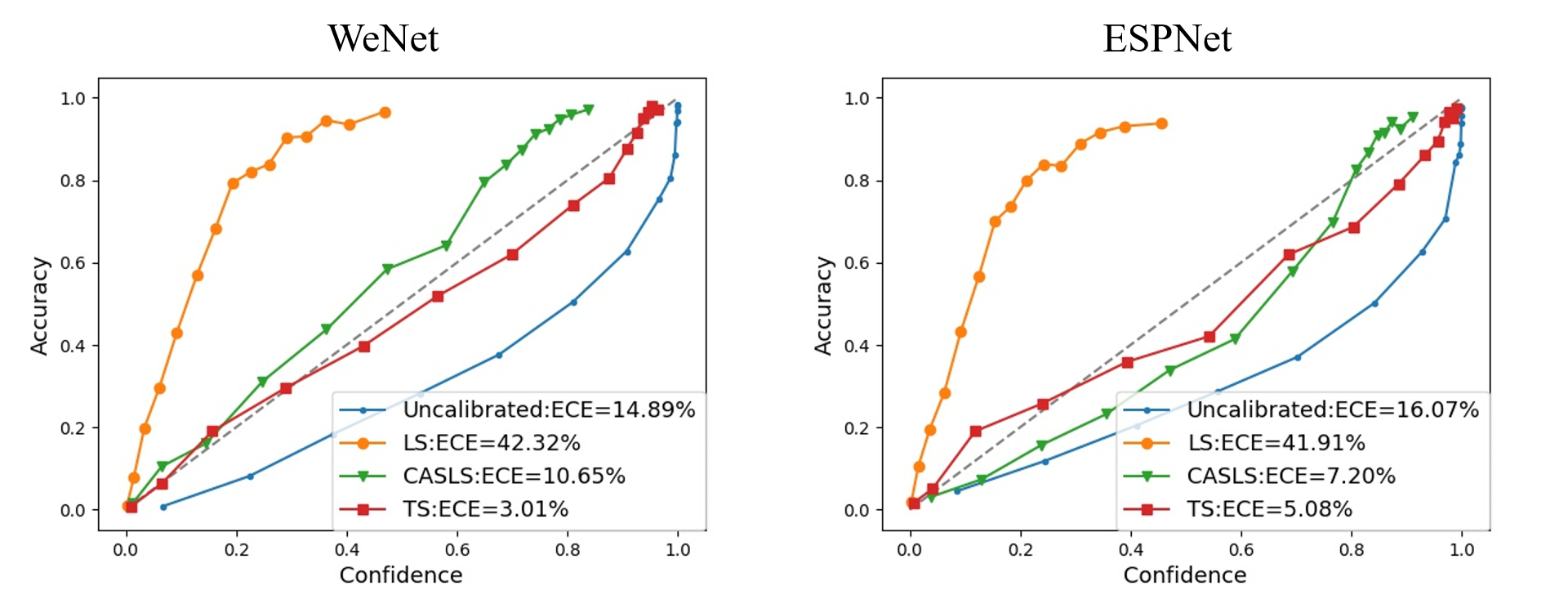}
  \caption{The reliability diagrams (with $B_m$=15 bins) of the uncalibrated models and the models calibrated with LS, TS and CASLS on SR task.}
\label{figure6}
\end{figure}

\begin{table*}[ht]
\caption{The calibration results comparison of LS, TS and CASLS on SR task. The four metrics: accuracy(Acc), WER, ECE and brier score(BS) are listed.}
\label{table2}
\begin{tabular}{ccccccccc}
\hline
\multirow{2}{*}{Method} & \multicolumn{4}{c}{WeNet}    & \multicolumn{4}{c}{ESPNet} \\ \cline{2-9} 
 & Acc(\%) $\uparrow$ & WER(\%) $\downarrow$  & ECE(\%) $\downarrow$   & BS(\%)  $\downarrow$    
 & Acc(\%) $\uparrow$ & WER(\%) $\downarrow$  & ECE(\%) $\downarrow$   & BS(\%)  $\downarrow$                                                  \\ \hline
Uncalibrated            & 62.01   & 5.75    & \begin{tabular}[c]{cc}14.89\\ (+0\%)\end{tabular}   & \begin{tabular}[c]{@{}c@{}}15.18\\ (+0\%)\end{tabular}   & 62.11   & 5.13    & \begin{tabular}[c]{@{}c@{}}16.07\\ (+0\%)\end{tabular}   & \begin{tabular}[c]{@{}c@{}}16.62\\ (+0\%)\end{tabular}   \\ \hline
LS                      & 62.56   & 5.30    & \begin{tabular}[c]{@{}c@{}}42.32\\ (+184\%)\end{tabular} & \begin{tabular}[c]{@{}c@{}}34.74\\ (+129\%)\end{tabular} & 61.27   & 5.30    & \begin{tabular}[c]{@{}c@{}}41.91\\ (+160\%)\end{tabular} & \begin{tabular}[c]{@{}c@{}}35.33\\ (+113\%)\end{tabular} \\ \hline
TS                      & 62.00   & 5.57    & \begin{tabular}[c]{@{}c@{}}3.01\\ (-80\%)\end{tabular} & \begin{tabular}[c]{@{}c@{}}12.20\\ (-20\%)\end{tabular} & 59.94   & 5.68    & \begin{tabular}[c]{@{}c@{}}5.08\\ (-68\%)\end{tabular} & \begin{tabular}[c]{@{}c@{}}12.77\\ (-23\%)\end{tabular} \\ \hline
CASLS                   & 63.77   & 5.25    & \begin{tabular}[c]{@{}c@{}}7.00\\ (-53\%)\end{tabular}   & \begin{tabular}[c]{@{}c@{}}12.57\\ (-17\%)\end{tabular}  & 64.08   & 4.72    & \begin{tabular}[c]{@{}c@{}}7.20\\ (-55\%)\end{tabular}   & \begin{tabular}[c]{@{}c@{}}13.43\\ (-19\%)\end{tabular}   \\ \hline
\end{tabular}
\end{table*}

\subsection{Speech Recognition}
\subsubsection{Experiment setup}

We conduct the experiment on AISHELL-1 corpus \cite{BuDNWZ17}. It is a large-scale Mandarin benchmark that contains about 170 hours speech data recorded from 400 speakers. And there are a total of 141,600 sentences in the recording. In the corpus, the training dataset contains 120,098 sentences from 340 speakers, the test dataset contains 7,176 sentences from 20 speakers. The remaining data valid dataset is utilized as the support dataset to obtain the general prediction distribution.

And we adopted the recent state-of-the-art toolkits as baseline: the WeNet proposed by Gulati et al. \cite{abs-2102-01547} and the ESPNet proposed by Watanabe et al.\cite{2018ESPnet}. The WeNet utilizes a shared encoder that consists of multiple Transformer\cite{VaswaniSPUJGKP17} or Conformer \cite{GulatiQCPZYHWZW20} encoder layers to extract the phonetic feature. And a joint CTC and attention decoder is used for sequence prediction. And ESPNet utilizes recurrent neural network as encoder, and adopts the same decoder as WeNet does. It is noted that both models are trained with LS ($\alpha^{'}$=0.1) already. For a fair comparison, we retrain the models with cross-entropy loss to serve as uncalibrated models. And the CASLS is conducted based on the uncalibrated ones.

During the calibration, we reduce the dimensions of confusion matrix as the number of phenom tokens is too large (4,000 classes) to calculate in the limited computer memory. Specifically, only the classes that belong to the error-prone set are reserved in the confusion matrix. Moreover, not only the prediction of many classes is sparse, but also the correct rate of the corresponding classes is extremely low. Therefore, the threshold of the error-prone $T$ should be low. Here, we set up the threshold of the error-prone $T$=0, the smoothing parameter $\alpha$=0.2, and the interval width of each bin $B_m$=15.

\subsubsection{Results and Discussion}

Fig. \ref{figure6} shows the reliability diagram of WeNet and ESPNet. Similarly, all the uncalibrated models (blue curves) and the models calibrated by the LS (orange curves) suffer from the over-confident and less-confident problems, respectively. The highest sequential confidence is even lower than 0.6 after conducting LS on each token on both models. And the TS (red curves) and CASLS (green curves) approaches can achieve a relatively better calibration performance.

In table \ref{table2}, the calibration performance of LS and CASLS are listed. And an additional metric, word error rate (WER), is added based on the metrics used in STR task. Both models trained with CASLS achieve the improvement in accuracy and the reduction in WER. In terms of ECE and BS, the CASLS reduce the ECE by 53\% on WeNet. On ESPNet, the effect of CASLS is slighter, but still achieves a 55\% reduction in ECE. Although TS performs best in terms of ECE and BS, our proposed method can effectively improve the recognition metrics with the second best performance on calibration metrics.

\section{Conclusion}
{
In this paper, we have investigated the problem of confidence calibration for sequence recognition. A so-called Context-Aware Selective Label Smoothing (CASLS) method has been proposed, which boosts label smoothing by leveraging the intrinsic contextual dependency underlying sequences and the statistical priors of class-specific prediction errors. Intensive experiments for two sequence recognition tasks, i.e., scene text recognition and speech recognition, both on publicly available benchmarking datasets, have been carried out to demonstrate our CASLS can achieve state-of-the-art performance. Ablation studies have also been conducted to further verify the effectiveness of the proposed method. In our future work, we will further verify our method on more sequential recognition tasks. We will also explore more effective strategies for making use of the contextual information.}

\section{Acknowledgements}
{
The research is supported in part by Guangdong Basic and Applied Basic Research Foundation (No. 2021A1515012282), National Nature Science
Foundation of China (Granted No. 61936003, 62076099) and the Alibaba Innovative Research (AIR) Program.}

\vfill\eject
\bibliographystyle{ACM-Reference-Format}
\balance
\bibliography{sample-base}

\end{document}